# Knowledge-based Drug Samples' Comparison

GUILLEMIN Sébastien
*Laboratoire d'Informatique de Bourgogne (LIB EA 7534)*
*University of Burgundy*
Dijon, France
sebastien.guillemin@u-bourgogne.fr

ROXIN Ana
*Laboratoire d'Informatique de Bourgogne (LIB EA 7534)*
*University of Burgundy*
Dijon, France
0000-0001-9841-0494

DUJOURDY Laurence
*Cellule d'Appui à la recherche en science des données*
*Institut Agro Dijon*
Dijon, France
laurence.dujourdy@agro-dijon.fr

JOURNAUX Ludovic
*Laboratoire d'Informatique de Bourgogne (LIB EA 7534)*
*Institut Agro Dijon*
Dijon, France
ludovic.journaux@agro-dijon.fr

*Abstract — Drug sample comparison is a process used by the French National Police to identify drug distribution networks. The current approach is based on a manual comparison done by forensic experts. In this article, we present our approach to acquire, formalise, and specify expert knowledge to improve the current process. We use an ontology coupled with logical rules to model the underlying knowledge. The different steps of our approach are designed to be reused in other application domains. The results obtained are explainable making them usable by experts in different fields.*

## I. INTRODUCTION

The fight against drug trafficking has been one of the French government's priorities since the end of 2019 and has led to the creation of the National *Stup* plan. This plan comprises 55 measures, including the use of new indicators to understand consumer habits and dealers' methods. The work described in this article is part of this plan and aims to support scientific experts in the decision-making process for narcotic profiling.

As part of the fight against drug trafficking, several arrests may be made, often accompanied by seizures. Forensic experts perform several analyses on samples from a seizure. They aim to correlate different samples from different seizures to identify trafficking networks best. To do so, experts use sample matching to pair samples according to their characteristics. Paired samples constitute an ensemble called a *batch*. The sample characteristics used are represented by different data, namely: macroscopic data (e.g., sample dimension, drug logos), qualitative data (e.g., list of active substances), quantitative data (e.g., dosage of substances) or non-confidential seizure data (e.g., date, place of seizure). This data concerns cannabis samples and tablets such as amphetamine. In France, such data is stored in the national STUPS© database.

In this context, we present an approach for modelling both business domain knowledge and the analysis rules used in that domain. Our approach also covers the application of these rules to real data. The purpose of this work is to be used in a decision-making support process for reducing the workload of experts.

This article is structured as follows: Section II presents the scientific background and definitions needed to understand this article. Section III presents the related works. Sections IV and 0 present our approach and the results we obtained by applying it to drug samples' comparison. Section 0 discusses our approach compared to the works exposed in section III. Finally, section VII presents our conclusions along with future works.

## II. SCIENTIFIC BACKGROUND AND DEFINITIONS

### A. Symbolic artificial intelligence

Symbolic Artificial Intelligence (AI) aims to represent and reproduce human cognitive reasoning using symbols in the form of knowledge representation systems [1]. Symbolic AI requires formal and explicit representations of a knowledge domain and mechanisms for deducing implicit knowledge from explicit facts. To do so, different logical languages can be used. We will introduce Description Logics (DL) [2] as it is used in our approach.

#### 1) Description Logics

DLs are a subset of First Order Logic (FOL) [3]. Unlike FOL, DLs are generally decidable. In other words, specific algorithms (called decision procedures) can be applied over DL rules and will (generally) return a result in a finite time. DLs are used for knowledge representations in specific domains, primarily because of the adaptability of their expressiveness and their overall decidability.

In DL, languages are characterised by a set of constructors. Constructor combinations determine a language's expressivity. The more constructors are used, the more expressive a language becomes. However, raising expressiveness increases the execution time of decision procedures.

The basic (i.e., the minimal) description language is the Attributive Language with Complement ($\mathcal{AL} - \mathcal{C}$) language [2]. With this language concept descriptions are specified using atomic concepts as unary predicates and atomic roles as binary predicates. Using this language, concepts can be defined as:

- the universal concept ($\top$)
- the bottom concept ($\bot$)

- the atomic negation of an atomic concept $A$ ($\neg A$)
- the intersection of two concepts $C$ and $D$ ($C \sqcap D$)
- a concept with value restriction ($\forall R.C$)
- a concept with limited existential quantification ($\exists R.\top$)
- the complement of a concept $C$ ($\neg C$)

Starting from the $\mathcal{AL-C}$ language, additional constructors can be added to form more expressive languages. For example, by using the *union* constructor (noted $\sqcup$) a concept C can be defined as equivalent to the union of two other concepts, A and B: $C \equiv A \sqcup B$. The complete list of available constructors is out of the scope of this article and is provided in [2].

*2) Ontology*

An ontology is an explicit and formal specification of a shared conceptualisation of a knowledge domain [4]. Ontologies are expressed using DL [1] languages and used for knowledge representation. An ontology comprises two main parts: the *Terminological Box* (TBox) and the *Assertionnal Box* (ABox). The TBox describes the terminological knowledge, i.e., the concepts and their properties. The ABox contains the instances of the concepts of the concept described in the TBox. A knowledge box (KB) is the combination of a TBox and an ABox.

An ontology is made to be easily sharable and reusable. This is a very interesting characteristic because it reduces the work required to describe the related domain knowledge.

*3) Reasoner*

A reasoner [5] is an algorithm that can validate and enrich ontology knowledge at a TBox or an ABox level. As an example, consider three concepts *C1*, *C2* and *C3* defined in the TBox. If the transitive relation *r* links *C1* to *C2* and, *C2* to *C3* a reasoner will infer that *C1* is also linked to *C3* by *r* due to its transitive characteristic.

Implementing deduction processes is based on decision procedures, i.e., algorithms complying with the following 3 properties:

- **Stop**: The algorithm must give the result in a finite time.
- **Correctness**: The inferences produced are consistent with the associated semantics, meaning that what is syntactically true is also semantically true.
- **Completeness**: All valid formulas can be demonstrated on the syntactic level.

One of the most used decision procedures is the Tableaux algorithm adapted for DL [6]. It is based on rebuttal evidence, i.e., a logical formula is verified by demonstrating that its negation is a contradiction.

*B. Semantic Web*

The Semantic Web (SW) is *"a vision for the future of the Web in which information is given explicit meaning"* [7]. It is an ensemble of standards and technologies for defining computer processable knowledge representations. A complete description of the SW technologies is out of the scope of this article. The sections below provide a short overview of those used in our approach.

*1) Ontology in Semantic Web*

*a) Web Ontology Language*

Web Ontology Language (OWL) families are ontology description languages based on DL that extend RDFS [8]. OWL 2 [9], the last version of OWL, defines three profiles enabling users to take advantage of certain features. Each profile has its own expressivity. From the least to the most expressive, these profiles are:

- OWL2 EL: allows subclass axioms with the intersection, existential quantifier, all, nothing, and closed classes with a single member. It does not support negation, disjunction, universal quantifier, or inverse properties.
- OWL2 QL: allows sub-properties, definition of sub-classes and domains/scopes. It does not support closed classes.
- OWL2 RL: allows all axiom types, cardinality restrictions (on scope only ≤1 and ≤0), and closed classes with a single member. It does not allow some constructors (universal quantifier and negation on domain, existential quantifier and union of classes for scope).

*b) TBox and ABox*

Ontology TBox explicitly describes concepts of a particular domain using classes. A class indicates the necessary and sufficient conditions for an instance to belong to this class. Each class may have a set of data properties representing a concept's features. For example, a sample of cannabis may have a weight and/or a height feature. Relations between concepts are represented by object properties between the corresponding classes.

Besides the TBox, the ABox contains class instances (also called *individuals*). An instance can be seen as an object belonging to a class. To belong to a class, an instance must respect conditions defined in the TBox for this class. In this object, data properties have real values (from real data), and object properties link the instances together.

*c) Reasoners*

Different reasoners are used in SW. For example, Pellet [10] or RacerPro [11] are based on the Tableaux algorithm. They are both open-source reasoners based on OWL-DL.

The different DL constructors supported by these reasoners are different. On the first hand, Pellet supports $\mathcal{SROIQ}(D)$ constructors and, on the other hand, RacerPro supports $\mathcal{SHIQ}$ constructors.

*2) SPARQL*

SPARQL Protocol and RDF Query Language (SPARQL) [12] is a set of specifications that provide languages and protocols to query and manipulate RDF [13] graphs. SPARQL syntax is made of different clauses to refine queries. In particular, the *WHERE* clause is used to query data according to a certain condition. For example, the following query is used to find mothers with at least one child over the age of 10:

```
SELECT DISTINCT ?mother
WHERE
{
  ?mother a :Mother.
  ?mother :hasChild ?child .
  ?child :age ?age

  FILTER(?age > 10)
}
```

A SPARQL variable starts with a question mark. The above example uses three variables: *?mother*, *?child* and *?age*. The *FILTER* clause is used to filter out all children under the age of 10.

*C. Domain Knowledge and Analysis Rules*

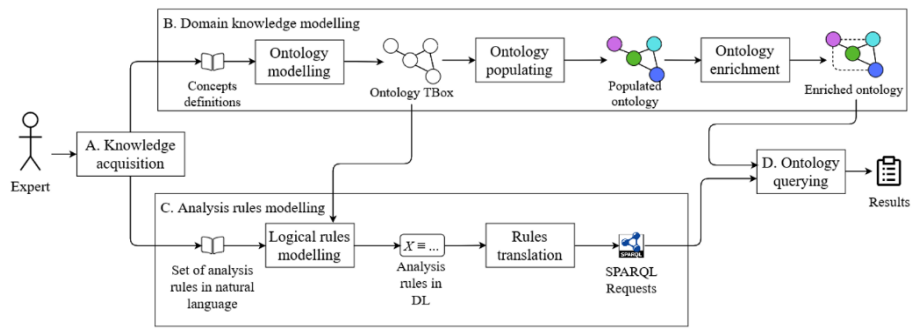

*Figure 1: Steps from knowledge acquisition to ontology querying.*

The approach described in section IV required experts' knowledge. We consider this knowledge as being composed of two parts: the *Domain Knowledge* and the *Analysis Rules*. The *Domain Knowledge* comprises our KB namely the descriptive knowledge of the domain i.e., the concepts of the domain, their relations, and their features (the TBox), along with concepts' instances (the ABox). For example, *Domain Knowledge* contains the statements: the *Sample* concept is part of the *Seizure* concept of and has several macroscopic features (modelled as object properties such as *weight*, *height* etc.). It also contains several instances of these concepts.

The component *Analysis Rules* contains the rules corresponding to the reasoning process used by the experts to deduce facts. It therefore completes the domain knowledge. For example, if two samples have the same features and are seized on close dates, then it is possible to conclude that these samples come from the same dealer.

III. RELATED WORK

*A. Forensic science*

Analysis of drug data has been the object of several works. Our work relates to drug distribution network detection and drug profiling. This domain aims to match drug samples according to their characteristics to identify distribution networks. Different studies were conducted according to the drug type, for example, heroin ([14], [15], [16], [17], [18] and [19]), cocaine ([16], [17], [18]), MDMA or fentanyl ([20], [21] and [22]). Other works are more general and review the different methods used in forensics ([23] and [24]).

The evolution of trends has also been the subject of various studies ([25] and [26]). This work seeks to analyse how different substances have been used by drug producers over the years. The results of these studies are useful for making assumptions about future trends. This helps the authorities to consider proactive measures.

Finally, there are publications on the use of AI in forensic science, field in which AI is an assisting tool attempting to overcome the limits of human biases in traditional approaches. Different uses of AI are reviewed in [27]. [28] focuses on the benefits and limitations of the use of AI methods in the forensic field.

To the best of our knowledge, only statistical approaches have been used to process the data and assist the forensic experts. No previous approach considers formal and explicit experts' knowledge modelling while comparing drug samples.

*B. Knowledge Graph Completion*

Knowledge Graph Completion (KGC) is a field that deals with enriching incomplete knowledge graphs [29]. This field is linked to our approach in the sense that the expert analysis rules are used precisely to complete knowledge of the domain. Many approaches based on statistical methods exist ([29] and [30]). For example, R-GCN [31] is based on messages passing between neighbour nodes according to the different relations in the graph. More traditional approaches can also be considered. [30] divided traditional approaches into heuristic, latent-feature, and content-based approaches. With statistical approaches, results are challenging to explain, making them unusable in critical domains such as medical or legal fields.

To the best of our knowledge, only a few methods based on a purely symbolic approach exist. For our approach, we have taken inspiration from such an existing method [32]. In this approach, SWRL [33] rules are used to enrich the knowledge base. The disadvantage of this approach is that links are automatically added to the knowledge base, leaving the expert no choice.

## IV. OUR APPROACH

Our approach comprises four main steps: *knowledge acquisition*, *domain knowledge modelling*, *analysis rules modelling* and *ontology querying*. Each step is described in the following sections. The overall process is illustrated in Figure 1.

*A. Knowledge acquisition*

The knowledge acquisition consists in acquiring the required knowledge of the application domain. By exchanging with the domain experts, the knowledge engineer builds two corpuses (i.e., natural language descriptions), one for each part of the domain knowledge (*Domain knowledge* and *Analysis rules*). Hence, he must obtain a set of concept definitions and a set of analysis rules (both in natural language).

This *Knowledge Acquisition* step is subject to the *knowledge acquisition bottleneck* [34]. It refers to the problem of slow and inaccurate knowledge acquisition while exchanging with the experts. Several factors impact the knowledge bottleneck, such as difficulties in understanding the business domain or lack of expert cooperation. Therefore, to facilitate exchanges with the experts, it is a good practice to follow a methodology such as the one described in [35]. This kind of methodology helps to identify key concepts and relations that will be used when modelling the ontology (see IV.B.1). For example, the knowledge engineer can seek to answer the following questions:

- What queries must the ontology allow answering?
- Who will use and maintain the ontology?
- What concepts are used, and what is their definition?
- What are the necessary and sufficient conditions for an instance to belong to a class?

Additionally, analysing databases allows for identifying additional knowledge and constraints. For example, the names of the tables of a relational database can help to identify certain concepts. In the same way, the presence of foreign keys can help to find concept relations. The conclusions drawn from the database analysis must always be validated by the experts. We assume that the knowledge engineer is not able to judge the relevance of the discovered knowledge by himself.

*B. Domain knowledge modelling*

The second step deals with modelling the knowledge according to the definitions obtained previously. This step is divided into three sub-steps: *Ontology modelling*, *Ontology populating* and *Ontology enrichment*.

*1) Ontology modelling*

The first sub-step is called *Ontology modelling*. It consists of modelling the ontology TBox according to the definition obtained during the *Knowledge acquisition* step (see IV.A). The process is the following:

1. For each concept, create a class in the ontology TBox having the same name as the concept (e.g., *Person*).
2. For each class, identify the concept properties in the associated concept definition. For each of these properties, create a data property whose domain is the class associated with the concept and whose range is the property's data type (e.g., xsd:string). The name of the new data property is the name of the concept property (e.g., class *Person* may have the data property *age*).
3. From each definition, identify relations between concepts. For each of these relations, create an object property and set its domain and range by the definitions. The name of the object property depends on the nature of the relation between the two concepts. Generally, the name of object properties starts with a verb like "*has…*" or "*is…*" (e.g., *hasFather*).

It is a common practice to create an inverse object property for each object property (some applications may require not to create it). This inverse object property will have a name beginning with an inverse verb (e.g. *has* becomes *is..of* i.e. *hasFather* becomes *isFatherOf*) and its domain (respectively its range) corresponds to the range (respectively the domain) of the starting property.

Another good practice is to add a comment describing the concept for each TBox class. To do so, the knowledge engineer can use the descriptions of the concepts obtained after the *Knowledge acquisition* step. This comment can then avoid ambiguity when the ontology is used by another user.

*2) Ontology population*

The next sub-step, *Ontology populating*, consists in instantiating the different TBox classes with data. The technical processes used to carry out this task depends on the application and the nature of the data. It is then difficult (even counter-productive) to give a precise action plan for this step. However, we can give you a summary of the main processes involved.

The first process is to retrieve the useful data from the different databases. To do this, the knowledge engineer must explore the data, possibly with the help of an expert who is familiar with the data. The notion of usefulness implies using only required data for concept instantiations. Handling only required data reduces the workload in the next processes and the risk of errors when instantiating concepts. Naturally, this data may be different if the project evolves and the TBox needs to be modified.

The second process consists of transforming the data to fit the TBox description. This process is particularly useful for switching from one datatype to another or to clean the data. Consider a trivial case where an ontology class has a data property representing a date (e.g., xsd:datetime) but this data is stored as a string in the database. In this case, the transformation process will convert the data from a string to an xsd:datetime format. Another example is when special characters are present in strings (such as accents). If these characters are not allowed (for some reason) then they are removed during this process.

The final process aims to use the transformed data to instantiate the TBox. Triplets will then be added to the original graph to form the ABox. It is also during this step that the graph can be loaded in a triple store that supports SPARQL querying.

*3) Ontology enrichment*

The final sub-step is the *Ontology enrichment*. It uses the populated ontology obtained from the previous sub-step. The knowledge described by this ontology may be incomplete. This may be due either to a lack of understanding of the domain by the knowledge engineer or to their choice not to describe all the knowledge. It may make sense not to describe all the knowledge explicitly to simplify the previous step. For example, if an object property is transitive, then it is not necessary to specify this property between all the instances of the concepts concerned. The ontology enrichment stage will add the missing properties.

This enrichment is achieved by using a reasoner to infer new knowledge. The choice of reasoner is therefore important. It must support all the semantics of the language chosen to describe the ontology. For example, if the language chosen is OWL-RL, the reasoner chosen must support the restriction on cardinalities.

*C. Analysis rules modelling*

The *Analysis rules modelling* step consists of using the analysis rules obtained from the *Knowledge Acquisition* step to create SPARQL requests (see section IV.C.2). These requests will then be used to query the ontology.

This step is divided into two sub-steps: *Logical rules modelling* and *Rules translation*.

*1) Logical rules' modelling*

This first sub-step aims to create logical rules from the natural language description of the analysis rules. The purpose of this step is to ensure that the rules do not contradict each other. Using a logical formalism eliminates any ambiguity that might be caused by descriptions in natural language.

The logical rules are written using DL (see section II.A.1). The set of constructors must then be chosen to be large enough to model all the analysis rules correctly, but not too large because it would impact the reasoning time. When finding the right set of constructors before modelling is complex, we suggest starting with a large set and then reducing it, once the rules have been modelled, to keep only the essential constructors. This step also involves using a reasoner to verify whether the rule set is consistent. As for the *Ontology enrichment* step (see IV.B.3) the reasoner must be chosen by the chosen constructors.

To model logical rules, the knowledge engineer must identify the elements involved in the natural language formulation of the rules (i.e., concepts, object properties, data properties etc.). These elements must necessarily be part of the ontology TBox. Indeed, as the SPARQL requests obtained from the logical rules will be used to query the ontology all the elements used in the logical rules must be correctly defined in the ontology. Therefore, the TBox must be revised if it is impossible to match an element of an analysis rule with an element of the ontology (i.e., the *Knowledge acquisition* and *Ontology modelling* steps must be redone).

The different identified elements are then translated into DL rules. To do so, concepts are translated into unary predicates and data properties and object properties are translated into binary predicates.

*2) Rules translation*

The final sub-step of *Analysis rule modelling* is the *Rules translation*. The logical rules obtained from the previous sub-step are translated into SPARQL queries. These queries will be used by the ontology querying step (see section IV.D).

As queries are used in the context of decision support, they must not directly modify the ontology. Instead, each query must indicate whether the rule is satisfied or not. A possible way to do this is to return a Boolean indicating if the rule is satisfied.

As an example, consider a DL rule stating that two persons are siblings if they have the same father:

$$Siblings(p2, p3) \equiv Person(p1) \cap Person(p2) \cap Person(p3) \cap isFatherOf(p1, p2) \cap isFatherOf(p1, p3)$$

The following SPARQL Query can then be obtained from this rule (we do not specify the prefix to simplify reading):

```
SELECT ?p2 ?p3 ?siblings
WHERE
{
  ?p1 a :Person .
  ?p2 a :Person .
  ?p3 a :Person .

  ?p1 :isFatherOf ?p2 .
  ?p1 :isFatherOf ?p3 .

  BIND((?p2 != ?p3) as ?siblings)
}
```

The above query selects all the pairs of instances of the *Person* class that have the same father and indicates if they are siblings. This query can be improved, for example by excluding all cases where *?p1* and *?p2* refer to the same instance. In this case, only the siblings would be displayed and therefore there would be no need to use the *?siblings* variable. This example shows that the selected information in the query depends on the experts' needs to make a decision.

The knowledge engineer can follow a guideline to construct the *WHERE* clause of the query. Each unary predicate *P(x)* of the DL rule is translated in a statement *?x a :P*. Each binary predicate of the form *B(x, y)* is translated in a statement *?x :B ?y*. A *BIND* clause can be used inside the *WHERE* clause to set the value of a Boolean variable.

*D. Ontology querying*

The final step of our approach is *Ontology Querying*. It consists of querying the ontology with the previously obtained SPARQL requests. The result of each request is presented to the experts to help them decide. Results must be clearly displayed, i.e., an expert must understand intuitively which result belongs to which request. Result presentation is thus crucial as it must highlight only the relevant elements for the experts.

V. RESULTS

In this section, we present the results we obtained by using our approach in the context of the *Stup* plan for *semi-automatic matching of drug samples* (see section I).

*A. Knowledge acquisition*

The domain knowledge acquired by exchanging with the experts is composed of 20 concept definitions. As an example, a *Sample* is defined by the experts as follows:

*A drug sample is extracted from a sealed sample. A sample is characterised by a sample number and its drug type ("cannabis", "cocaine", "miscellaneous" or "amphetamine and derivatives"). A sample has macroscopic characteristics (internal and external appearance, height, width etc.), active principles and cutting products. Experts can also comment on the sample. In addition, a sample may be grouped with other samples in a batch. In the case of narcotics on which a chemical profiling analysis is carried out, each sample is associated with a chemical profile.*

We show in section V.B how this description and the STUPS© data are used to obtain a populated ontology.

Concerning the analysis rules, a set of nine rules has been obtained describing the conditions that two samples must have to be matched. Here is one of these rules:

*Two samples match if their drug types are the same.*

We show in section V.C how this rule is then translated into a DL rule and a SPARQL request.

*B. Domain knowledge modelling*

The first step to model the domain knowledge is the *Ontology modelling* step. To do so, we used the domain knowledge expressed in natural language obtained previously. Using the above definition of a sample, we specify the following object properties for the *Sample* concept:

- hasExternalAspect (range: *Aspect*)
- hasInternalAspect (range: *Aspect*)
- hasActivePrincipal (range: *ActivePrincipale*)
- hasCuttingProduct (range: *CuttingProduct*)
- hasChimicalProfile (range: *ChimicalProfile*)
- isCloseTo (range: *Sample*)
- comesFrom (range: *Sealed*)

Additionally, the following data properties are added to the *Sample* concept:

- sampleNumber (rdfs:range *xsd:string*)
- drugType (rdfs:range: *{"cannabis", "cocaine", "miscellaneous", "amphetamine and derivatives"}*)
- comment (rdfs:range *xsd:string*)
- height (rdfs:range *xsd:float*)
- width (rdfs:range *xsd:float*)
- diameter (rdfs:range *xsd:float*)
- thickness (rdfs:range *xsd:float*)
- length (rdfs:range *xsd:float*)

The Sample concept contains other data properties corresponding to other macroscopic characteristics which are not provided here but can be found in the ontology TBox provided below.

Using the definition of all the other concepts, we modelled an ontology TBox made of 20 concepts, 45 object properties and 40 data properties. Each concept is annotated with its definition. This ontology TBox can be found on GitHub (github.com/SebastienGuillemin/StupsOntology). Our ontology complexity is SROIQ(D) [36]:

- S is an abbreviation for ALC.
- R refers to the use of roles.
- O refers to value restriction (*owl:oneOf*)
- I is used for inverse properties.
- Q is used for cardinality restrictions.
- (D) refers to the use of datatype properties.

After modelling the ontology, we performed the *Ontology populating* step. This step uses data from the STUPS© to instantiate the ontology TBox. Explaining the program to retrieve and transform the data is out of the scope of this article but it is available on GitHub (github.com/SebastienGuillemin/etl). When populating the ontology, 68,972 instances were created, 20,001 of which were *Sample* instances. Figure 2 shows an example of a *Sample* instance and its relations with other instances.

To perform the *Ontology enrichment*, we loaded the populated ontology in a triplestore. We chose GraphDB which comes with several reasoners compatible with the complexity of our ontology. We chose the OWL-Max reasoner among the different reasoners available because it considers all the constraints used to define our TBox (graphdb.ontotext.com/documentation/10.3/owl-compliance.html). This Ontology enrichment added 382,205 new relations between instances in our ontology (increasing from 284,638 to 666,843). We have only measured the creation of relations concerning the data properties and object properties modelled in our ontology.

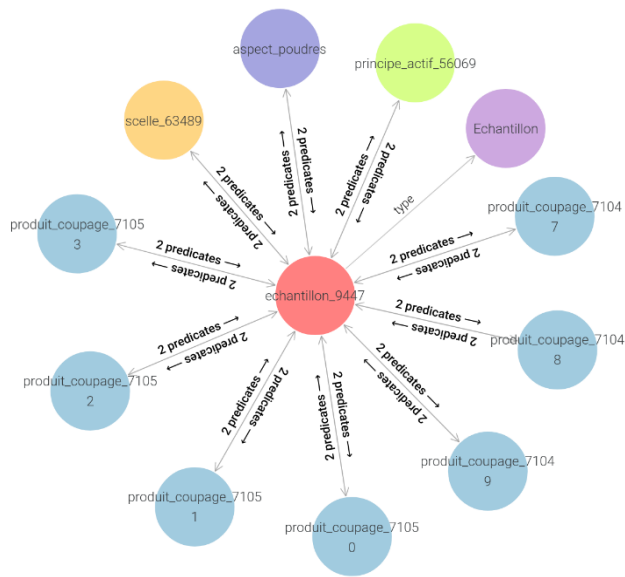

*Figure 2. Graph visualisation of a sample instance (in red). Cutting products are displayed in light blue, external aspect in dark blue, sealed in yellow and active principle in light green.*

C. *Analysis rules modelling*

The next step is to model the analysis rules. Using the knowledge acquired during the step described in V.A we will consider the following analysis rule: *Two samples match if their drug types are the same.*

We show how this rule is converted into a DL rule and then into a SPARQL request. From this analysis rule, we identify that the ontology elements to consider are the *Sample* class and its *drugType* data property. These elements are present in the previously obtained ontology TBox thus we can continue.

The resulting DL rule must be made of the unary predicate *Sample* and the binary predicate *drugType*. We obtain the following DL rule:

$$match(s1, s2) \equiv Sample(s1) \cap Sample(s2) \cap drugType(s1, dt1) \cap drugType(s2, dt2) \cap dt1 = dt2 \cap s1 \neq s2$$

Each rule of our corpuses of nine rules is converted into a DL rule in the same manner.

Once all the analysis rules have been modelled using DL, they are translated into SPARQL requests. The above DL rule is translated as follows (once again, we do not specify the prefix to simplify reading)

```
SELECT ?s1 ?s2 ?match
WHERE
{
  ?s1 a :Sample .
  ?s2 a :Sample .
  ?s1 :drugType ?dt1 .
  ?s2 :drugType ?dt2 .

  FILTER(?s1 != ?s2)
  BIND ((?dt1 = ?dt2) as ?match)
}
```

Here, the *?match* variable indicates whether *?s1* and *?s2* can be matched. We translated the DL unary predicate *Sample* by using the SPAQRL keyword "a". Thus, the two first lines of the **WHERE** clause ensure that *?s1* and *?s2* are *:Sample* instances. We then retrieve their respective drug type. To do so, we translate the binary predicate *drugType* into the relation *:drugType*. The drug type of *?s1* (respectively *?s2*) is bound to *?dt1* (respectively *?dt2*). Then, we ensure that *?s1* and *?s2* are not the same *:Sample* instance by using the **FILTER** clause. Finally, if *?dt1* and *?dt2* have the same value the *?match* variable is bound to *True*. Otherwise, *?match* is bound to *False*.

This translation process from DL rules to SPARQL requests is applied to all analysis rules. As we have 9 analysis rules, we end up with nine SPARQL requests.

*D. Ontology querying*

The final step of our approach consists of querying the ontology using the previously defined SPARQL request. We will illustrate this step by using the following analysis rules:

- *Two samples match if their drug types are the same.*
- *Two samples match if they have the same chemical form.*
- *Two samples match if their macroscopic features (i.e., height, width) differ by less than 5%.*

Each of these rules has been translated into DL and then into SPARQL. In order not to overload the article, we will not detail these steps for these rules.

We also consider the following samples:

|  | Sample 1 | Sample 1 |
|---|---|---|
| Sample number | 1 | 2 |
| Drug type | Cannabis | Cannabis |
| Chemical form | Resin | Resin |
| Width | 200 millimetres | 150 millimetres |
| Height | 100 millimetres | 100 millimetres |

Each SPARQL request of each rule is applied to these samples. This conduct to the following results :

- Two samples match if their drug types are the same: **True**.
- Two samples match if they have the same chemical form: **True**.
- Two samples match if their macroscopic features (i.e., height, width, etc.) differ less than 5 per cent.
  - Width: **False**.
  - Height: **True**.

Using these rules, experts can decide whether samples match or not. For example, in this case, they can consider that the width characteristic is not relevant and conclude that the two samples match.

## VI. Discussion

Our work proposes an approach to acquire expert knowledge and use it to help experts during their decision process. An important aspect of our approach is that it does not directly modify the ontology, leaving the experts to make decisions. To the best of our knowledge, our approach is the first to use expert knowledge in the sample-matching process. Other works were based on statistical approaches (as exposed in section II).

The overall process, from knowledge acquisition to analysis with rules, is independent of the application domain. Then, using it in other domains requiring analytical expertise is possible. Nonetheless, some steps may require the use of other methodologies. It is the case of TBox modelling, which is based on Noy and McGuinness's guide [35].

Results presented in section 0 show how our approach assists the experts. For each potential match (i.e., a pair of samples), the result of each rule is displayed to the author. Experts exploit the different results to decide.

So far, our approach is only limited to results' display. Moreover, our approach relies on SPARQL queries that can tend to slow down the overall process. Also, translating some analysis rules into SPARQL requests can be an arduous task. As part of our future work, we'll investigate replacing SPARQL queries with a simpler translation method.

## VII. Conclusion and future works

We presented an approach based on ontologies and logical rules for decision support in comparing drug samples. We have defined a general process composed of four main steps, each step being designed to ease its reproducibility.

Our approach has been successfully tested for assisting forensic experts in drug samples' comparison. Still, it has some limitations, that will be addressed in future works. Firstly, additional support to experts can be provided by allowing them to rank the different rules i.e., in the form of a hierarchy. Such ranking would enable controlling the impact (or the score) of each rule on the suggested decision. The most important rules would then have the greatest impact (or score) on the suggestions made by our approach.

Secondly, rather than using SPARQL queries, we could directly enrich the reasoner rule set with our logical rules obtained during *Logical rules' modelling* (see section IV.C.1). This would eliminate the need to execute queries one by one over the populated ontology. Thus, the whole process would be more fluid by removing one step. However, the time needed by the reasoner to make inferences could increase to a greater or lesser extent. This depends on the number of rules and their expressivity (i.e., the DL constructors chosen). Reducing the expressivity of the underlying TBox can help in limiting the reasoning overhead. Additionally, inferences produced by the reasoner must not automatically be materialised in the KB. The experts must be the only ones to decide what to add to the populated ontology. So, it will be necessary to provide a mechanism for experts to decide whether the matching should be added to the ontology.

Finally, matches are based solely on analysis rules which may not always be appropriate. In fact, in the actual process used by the experts, their intuition plays a very important role. Unfortunately, this intuition cannot be formalised in the analysis rules. One possible way of simulating expert intuition would be to use statistical approaches (e.g., neural networks). These statistical approaches could then be coupled with the approaches presented above to improve decision suggestions. Moreover, we can imagine that statistical models would allow suggesting new facts (e.g., additional relevant analysis rules) to the existing KB. This new knowledge would then be accepted or refuted by the domain experts. Approaches combining neural networks and symbolic AI belong to the domain of neuro-symbolic AI [37]. It seeks to benefit from the advantages of both areas without the disadvantages i.e., the learning capacity of neural networks and the explicability of symbolic AI.

## VIII. References

[1] M. Flasiński, *Introduction to Artificial Intelligence*. Cham: Springer International Publishing, 2016. doi: 10.1007/978-3-319-40022-8.

[2] F. Baader and W. Nutt, 'Basic Description Logics', in *The Description Logic Handbook*, F. Baader, D. Calvanese, D. L. McGuinness, D. Nardi, and P. F. Patel-Schneider, Eds., 2nd ed.Cambridge University Press, 2007, pp. 47–104. doi: 10.1017/CBO9780511711787.004.

[3] P. D. Magnus, *forall x: Calgary. An Introduction to Formal Logic*. [Online]. Available: https://philpapers.org/rec/MAGFXI.

[4] R. Studer, V. R. Benjamins, and D. Fensel, 'Knowledge engineering: Principles and methods', *Data & Knowledge Engineering*, vol. 25, no. 1–2, pp. 161–197, Mar. 1998, doi: 10.1016/S0169-023X(97)00056-6.

[5] C. M. Keet, 'An Introduction to Ontology Engineering', vol. 1.5. University of Cape Town, 2020.

[6] R. M. Smullyan, 'First-Order Logic. Preliminaries', in *First-Order Logic*, R. M. Smullyan, Ed., in Ergebnisse der Mathematik und ihrer Grenzgebiete. Berlin, Heidelberg: Springer, 1968, pp. 43–52. doi: 10.1007/978-3-642-86718-7_4.

[7] D. L. McGuinness and F. Van Harmelen, 'OWL web ontology language overview', in *W3C recommendation*, vol. 10.10, 2004.

[8] J. Broekstra and A. Kampman, 'Sesame: A generic Architecture for Storing and Querying RDF and RDF Schema', 2001.

[9] B., Motik, P.F., Patel-Schneider and B.C. Grau. 'OWL 2 Web Ontology Language Direct Semantics (Second Edition)' W3C Recommendation, 2012. Available at: https://www.w3.org/TR/owl2-direct-semantics/.

[10] E. Sirin, B. Parsia, B. C. Grau, A. Kalyanpur, and Y. Katz, 'Pellet: A practical OWL-DL reasoner', *Journal of Web Semantics*, vol. 5, no. 2, pp. 51–53, Jun. 2007, doi: 10.1016/j.websem.2007.03.004.

[11] V. Haarslev, K. Hidde, R. Möller, and M. Wessel, 'The RacerPro knowledge representation and reasoning system', *Semantic Web*, vol. 3, no. 3, pp. 267–277, 2012, doi: 10.3233/SW-2011-0032.

[12] World Wide Web Consortium, 'SPARQL 1.1 overview', 2013.

[13] P.J , Hayes and P.F., Patel-Schneider, 'RDF 1.1 Semantics', *W3C Recommendation 25*, 2014. Available at: https://www.w3.org/TR/rdf11-mt/.

[14] P. Esseiva, L. Dujourdy, F. Anglada, F. Taroni, and P. Margot, 'A methodology for illicit heroin seizures comparison in a drug intelligence perspective using large databases', *Forensic Science International*, vol. 132, no. 2, pp. 139–152, Mar. 2003, doi: 10.1016/S0379-0738(03)00010-0.

[15] V. Dufey, L. Dujourdy, F. Besacier, and H. Chaudron, 'A quick and automated method for profiling heroin samples for tactical intelligence purposes', *Forensic Science International*, vol. 169, no. 2, pp. 108–117, Jul. 2007, doi: 10.1016/j.forsciint.2006.08.003.

[16] F. Ratle, C. Gagné, A.-L. Terrettaz-Zufferey, M. Kanevski, P. Esseiva, and O. Ribaux, 'Advanced clustering methods for mining chemical databases in forensic science', *Chemometrics and Intelligent Laboratory Systems*, vol. 90, no. 2, pp. 123–131, Feb. 2008, doi: 10.1016/j.chemolab.2007.09.001.

[17] B. Le Daré, S. Allard, A. Couette, P.-M. Allard, I. Morel, and T. Gicquel, 'Comparison of Illicit Drug Seizures Products of Natural Origin Using a Molecular Networking Approach', *Int J Toxicol*, vol. 41, no. 2, pp. 108–114, Mar. 2022, doi: 10.1177/10915818211065161.

[18] P. Esseiva, L. Gaste, D. Alvarez, and F. Anglada, 'Illicit drug profiling, reflection on statistical comparisons', *Forensic Science International*, vol. 207, no. 1, pp. 27–34, Apr. 2011, doi: 10.1016/j.forsciint.2010.08.015.

[19] A.-L. Terrettaz-Zufferey, F. Ratle, O. Ribaux, P. Esseiva, and M. Kanevski, 'Pattern detection in forensic case data using graph theory: Application to heroin cutting agents', *Forensic Science International*, vol. 167, no. 2, pp. 242–246, Apr. 2007, doi: 10.1016/j.forsciint.2006.06.059.

[20] A. Bolck, C. Weyermann, L. Dujourdy, P. Esseiva, and J. van den Berg, 'Different likelihood ratio approaches to evaluate the strength of evidence of MDMA tablet comparisons', *Forensic Science International*, vol. 191, no. 1, pp. 42–51, Oct. 2009, doi: 10.1016/j.forsciint.2009.06.006.

[21] C. Weyermann *et al.*, 'Drug intelligence based on MDMA tablets data: I. Organic impurities profiling', *Forensic Science International*, vol. 177, no. 1, pp. 11–16, May 2008, doi: 10.1016/j.forsciint.2007.10.001.

[22] T. Cooman, C. E. Ott, and L. E. Arroyo, 'Evaluation and classification of fentanyl-related compounds using EC-SERS and machine learning', *Journal of Forensic Sciences*, vol. 68, no. 5, pp. 1520–1526, 2023, doi: 10.1111/1556-4029.15285.

[23] M. Bovens, B. Ahrens, I. Alberink, A. Nordgaard, T. Salonen, and S. Huhtala, 'Chemometrics in forensic chemistry — Part I: Implications to the forensic workflow', *Forensic Science International*, vol. 301, pp. 82–90, Aug. 2019, doi: 10.1016/j.forsciint.2019.05.030.

[24] A. Popovic, M. Morelato, C. Roux, and A. Beavis, 'Review of the most common chemometric techniques in illicit drug profiling', *Forensic Science International*, vol. 302, p. 109911, Sep. 2019, doi: 10.1016/j.forsciint.2019.109911.

[25] L. Dujourdy and F. Besacier, 'A study of cannabis potency in France over 25 years (1992–2016)', *Forensic Science International*, vol. 272, pp. 72–80, Mar. 2017, doi: 10.1016/j.forsciint.2017.01.007.

[26] I. Du Plessis *et al.*, 'Statistical evaluation of the IPSC ecstasy-database 1998-2002', presented at the FORENSIC SCIENCE INTERNATIONAL, CUSTOMER RELATIONS MANAGER, BAY 15, SHANNON INDUSTRIAL ESTATE CO, CLARE, IRELAND: ELSEVIER SCI IRELAND LTD, Sep. 2003, pp. 90–90.

[27] Y. C. Saw and A. K. Muda, 'An Overview Computational Intelligence Solution in Forensic Drug Analysis', 2016.

[28] N. Galante, R. Cotroneo, D. Furci, G. Lodetti, and M. B. Casali, 'Applications of artificial intelligence in forensic sciences: Current potential benefits, limitations and perspectives', *Int J Legal Med*, vol. 137, no. 2, pp. 445–458, Mar. 2023, doi: 10.1007/s00414-022-02928-5.

[29] Z. Chen, Y. Wang, B. Zhao, J. Cheng, X. Zhao, and Z. Duan, 'Knowledge Graph Completion: A Review', *IEEE Access*, vol. 8, pp. 192435–192456, 2020, doi: 10.1109/ACCESS.2020.3030076.

[30] M. Zhang, 'Graph Neural Networks: Link Prediction', in *Graph Neural Networks: Foundations, Frontiers, and Applications*, L. Wu, P. Cui, J. Pei, and L. Zhao, Eds., Singapore: Springer Nature Singapore, 2022, pp. 195–223. doi: 10.1007/978-981-16-6054-2_10.

[31] M. Schlichtkrull, T. N. Kipf, P. Bloem, R. van den Berg, I. Titov, and M. Welling, 'Modeling Relational Data with Graph Convolutional Networks'. arXiv, Oct. 26, 2017. Accessed: Jul. 31, 2023. [Online]. Available: http://arxiv.org/abs/1703.06103

[32] Y.-L. Chi, 'Rule-based ontological knowledge base for monitoring partners across supply networks', *Expert Systems with Applications*, vol. 37, no. 2, pp. 1400–1407, Mar. 2010, doi: 10.1016/j.eswa.2009.06.097.

[33] I. Horrocks, P.F. Patel-Schneider, H. Boley, S. Tabet, B. Grosof, and M. Dean., 'SWRL : A semantic web rule language combining OWL and RuleML', *W3C Member Submission*, vol. 21, no. 79, p. 1-31, Sept. 2004. Available at: https://www.w3.org/Submission/SWRL/.

[34] C. Wagner, 'Breaking the Knowledge Acquisition Bottleneck Through Conversational Knowledge Management', 2006.

[35] N. F. Noy and D. L. McGuinness, 'Ontology Development 101: A Guide to Creating Your First Ontology', 2001.

[36] I. Horrocks, O. Kutz, and U. Sattler, 'The Even More Irresistible SROIQ', Kr, 2006, pp. 57-57.

[37] P. Hitzler, A. Eberhart, M. Ebrahimi, M. K. Sarker, and L. Zhou, 'Neuro-symbolic approaches in artificial intelligence', *National Science Review*, vol. 9, no. 6, p. nwac035, Jun. 2022, doi: 10.1093/nsr/nwac035.